\documentclass[journal,onecolumn,draftcls,11pt]{IEEEtran}
\ifCLASSINFOpdf
\else
   \usepackage[dvipdfm]{graphicx}
\fi
\usepackage[cmex10]{amsmath}
\usepackage{amssymb,amsthm}

 \usepackage{algorithmic}
 \usepackage[tight,footnotesize]{subfigure}
 \newcommand{\etal}{{\it et al.}}
 \newcommand{\RR}{\mathbb{R}}

 \newcommand{\T}{{}^\top}

 \newcommand{\bm}[1]{\boldsymbol{#1}}

 \newcommand{\vb}{\boldsymbol{b}}

 \newcommand{\vq}{\boldsymbol{q}}

 \newcommand{\vy}{\boldsymbol{y}}
 \newcommand{\vw}{\boldsymbol{w}}
 
 \newcommand{\vv}{\boldsymbol{v}}
 \newcommand{\valpha}{\boldsymbol{\alpha}}

 \newcommand{\vxi}{\boldsymbol{\xi}}

 \newcommand{\mA}{\boldsymbol{A}}

 \newcommand{\mI}{\boldsymbol{I}}

 \newcommand{\minimize}{\mathop{\rm minimize}}
 \newcommand{\maximize}{\mathop{\rm maximize}}
 \newcommand{\argmin}{\mathop{\rm argmin}}
 
 \newcommand{\subjectto}{\mbox{\rm subject to}}

 \newcommand{\Eqref}[1]{Eq.~{\eqref{#1}}}
 \newcommand{\Eqsref}[1]{Eqs.~{\eqref{#1}}}
 \newcommand{\Secref}[1]{Sec.~{\ref{#1}}}
 \newcommand{\Figref}[1]{Fig.~{\ref{#1}}}

\begin{document}
 %
 \title{Dual Augmented Lagrangian Method for Efficient Sparse Reconstruction\footnote{This work has been submitted to the IEEE for possible publication. Copyright may be transferred without notice, after which this version may no longer be accessible.}}
 %
 %
 %

 \author{Ryota~Tomioka$^\dagger{}^\ast$\thanks{$^\ast$(Corresponding author) Department of Mathematical Informatics,
        University of Tokyo, Hongo 7-3-1, Bunkyo-ku, Tokyo 113-8656,
        Japan. TEL: +81-3-5841-6898, FAX: +81-3-5841-6897, E-mail: tomioka@mist.i.u-tokyo.ac.jp.} and
        Masashi~Sugiyama$^\dagger$\thanks{$^\dagger$ Department of Computer
        Science, Tokyo Institute of Technology, 2-12-1-W8-74, O-okayama, Meguro-ku, Tokyo 152-8552, Japan.}
 }

 %
 %

 \markboth{Submitted to IEEE Signal Processing Letters, 2009}%
 {Shell \MakeLowercase{\textit{et al.}}: }
 %



 \maketitle

 \begin{abstract}
 \boldmath
  We propose an efficient algorithm for sparse signal reconstruction
  problems. The proposed algorithm is an augmented Lagrangian method
  based on the dual sparse reconstruction problem. It is efficient when
  the number of unknown variables is much larger than the number of
  observations because of the dual formulation. Moreover, the primal
  variable is explicitly updated and the sparsity in the solution is exploited.
  Numerical comparison with the state-of-the-art algorithms shows
  that the proposed algorithm is favorable when the design matrix is
  poorly conditioned or dense and very large.
 \end{abstract}


 \noindent EDICS category: SAS-STAT, SAS-MALN

 %
 \IEEEpeerreviewmaketitle

 \section{Introduction}
 \IEEEPARstart{S}{parse} signal reconstruction has recently gained considerable
 interests in signal/image processing and machine learning. Sparsity is
 often a natural assumption in inverse problems, such as MEG/EEG source
 localization and image/signal deconvolution; sparsity enables us to identify
 a small number of active components even when the dimension is much larger
 than the number of observations. In addition, a sparse model is also
 valuable in predictive tasks because it can explain why it is able to
 predict in contrast to black-box models such as neural networks and
 support vector machines.

 In this paper we consider the following particular problem that
 typically arises in sparse reconstruction:
 \begin{flalign}
 \label{eq:form_bpdn}
{\rm (P)}&& &
  \minimize_{\vw\in\RR^n}\quad\frac{1}{2}\|\mA\vw-\vb\|^2+\lambda\|\vw\|_1,&&
 \end{flalign}
 where $\vw\in\RR^n$ is the coefficient vector to be estimated, $\mA\in\RR^{m\times n}$ is the design matrix, and $\vb\in\RR^m$ is
 the vector of observations. It is well known that $\ell_1$-norm penalty
 enforces $\vw$ to have many zero elements. It is called 
 lasso~\cite{Tib96} in the statistics, basis pursuit
 denoising~\cite{CheDonSau98} in the signal processing, and
 FOCUSS~\cite{GorRao97} in the brain imaging communities. 

 Various methods have been proposed to efficiently solve the
 optimization problem \eqref{eq:form_bpdn} (or its generalized
 versions).
 Iteratively reweighted shrinkage (IRS) is a popular approach for
 solving the problem~\eqref{eq:form_bpdn} (see \cite{GorRao97,CotRaoEngKre05,Bio06,PalWipKreRao06,FigBioNow07}).
The main idea of the IRS approach is to replace a
 non-differentiable (or non-convex) optimization problem by a series of
 differentiable convex ones; typically the regularizer (e.g., $\|\cdot\|_1$ in
 \Eqref{eq:form_bpdn}) is upper bounded by a weighted quadratic
 regularizer. Then one can use various existing algorithms to 
 minimize the upper bound. The upper bound is re-weighted after every
 minimization so that the solution eventually converges to the solution of
 the original problem~\eqref{eq:form_bpdn}.
 The challenge in the IRS framework is the {\em
 singularity}~\cite{FigBioNow07} around the coordinate axis. For example,
 in the $\ell_1$ problem in \Eqref{eq:form_bpdn}, any zero component
 $w_j=0$ in the 
 initial vector $\vw$ will remain zero after any number of
 iterations. Moreover, it is possible to create a situation that the
 convergence becomes arbitrarily slow for finite $|w_j|$ because the
 convergence in the $\ell_1$ case is only linear~\cite{GorRao97}. 
 Another recent work is the split Bregman iteration
 (SBI) method~\cite{GolOsh08}, which is derived from the Bregman
 iteration algorithm~\cite{YinOshGolDar08} in order to handle the noisy
 ($\lambda>0$) case. The Bregman iteration algorithm can be considered as an
 augmented Lagrangian (AL) method
 (see~\cite{Ber82,NocWri99,YinOshGolDar08}). By introducing
 an auxiliary variable $\tilde{\vw}$, the SBI approach decouples
 the minimization of the first and the second term in
 \Eqref{eq:form_bpdn}, which can then be handled independently. The two variables $\vw$ and
 $\tilde{\vw}$ are gradually enforced to coincide with each other.
Both IRS and SBI require solving a linear system of the size of the
number of unknown variables ($n$) repeatedly, which may become challenging
 when $n\gg m$.

 Kim~\etal~\cite{KimKohLusBoyGor07} developed an efficient
 interior-point (IP) method called l1\_ls. They proposed a truncated Newton method
 for solving the inner minimization that scales well when the
 design matrix $\mA$ is sparse. 

 The iterative shrinkage/thresholding (IST) (see
 \cite{FigNow03,DauDefMol04,ComWaj05,YinOshGolDar08}) is a classic method but
 it is still an area of active research~\cite{Nes07,WriNowFig09}. It 
 alternately computes the steepest descent direction on the loss term in
 \Eqref{eq:form_bpdn} and the {\em soft thresholding} related to the
 regularization term. 
 The IST
 method has the advantage that every iteration is extremely light (only
 computes gradient) and every intermediate solution is sparse. However the naive
 version of IST is sensitive to the selection of step-size. Recently
 several authors have proposed intelligent step-size selection
 criteria~\cite{Nes07,WriNowFig09}.

 In this paper we propose an efficient method that scales well when
 $n\gg m$, which we call the dual augmented Lagrangian (DAL).  
 It is an AL method similarly to SBI method but it is applied to the
 dual problem; thus the inner minimization is efficient when $n\gg m$.
 In addition, in contrast to the ``divide and conquer'' approach of SBI, 
 the inner minimization can be performed jointly over all the variables;
 it converges {\em super linearly} because the inner minimization
 is solved at sufficient precision (see~\cite{Ber82,NocWri99}).
Moreover, although the proposed method is based on the dual
 problem, the primal variable is explicitly updated in the computation as
 the Lagrangian multiplier. DAL computes soft thresholding after every
 iteration similarly to the IST approach but with an {\em improved
 direction} as well as an automatic step-size selection mechanism;
 typically the number of outer iterations is less than 10. The proposed
 approach can be applied to large scale problems with {\em dense} design
 matrices  because it exploits the sparsity in the coefficient vector
 $\vw$ in  contrast to the IP methods~\cite{KimKohLusBoyGor07}, which
 exploits the sparsity in the design matrix.

 This paper is organized as follows. In \Secref{sec:method}, the
DAL algorithm is presented; two  approaches for the inner minimization problem are discussed. 
 In \Secref{sec:results} we experimentally compare DAL to the state-of-the-art
 SpaRSA~\cite{WriNowFig09} and l1\_ls~\cite{KimKohLusBoyGor07}
 algorithms. We give a brief summary and future directions in 
 \Secref{sec:summary}. 

 \section{Method}
 \label{sec:method}
\subsection{Dual augmented Lagrangian method for sparse reconstruction}
\label{sec:dal}
Let $f(\vw)$ be the objective in \Eqref{eq:form_bpdn}. The challenge in
minimizing $f(\vw)$ arises from its non-differentiability. The proposed approach is based on the
minimization of a differentiable surrogate function $f_\eta(\vw)$.
In this section we derive the surrogate function $f_\eta(\vw)$ and
its gradient from the augmented Lagrangian function $L_\eta$ of the dual
problem of \Eqref{eq:form_bpdn}.

Using the Fenchel duality (see \cite[Sec.~5.4]{Ber99}) and a splitting
similar to SBI (in the dual), we obtain the
following dual problem of problem~\eqref{eq:form_bpdn} (see also \cite{ComWaj05}):
\begin{flalign}
\label{eq:dual_obj}
{\rm (D)} && &\maximize_{\vv\in\RR^n,\valpha\in\RR^m} & &
 -\frac{1}{2}\|\valpha-\vb\|_2^2+\frac{1}{2}\|\vb\|_2^2-\delta_\lambda^\infty(\vv), &&&&&&\\
\label{eq:dual_const}
&& &\subjectto & &\vv=\mA\T\valpha,
\end{flalign}
where 
$\delta_\lambda^\infty(\vv)$ is the indicator function~\cite{ComWaj05} of the
$\ell_\infty$ ball of radius $\lambda$, i.e.,
$\delta_\lambda^\infty(\vv)=0$ (if $\|\vv\|_\infty\leq\lambda$), and
$+\infty$ (otherwise).
It can be shown that the strong duality holds, i.e., the maximum of
\Eqref{eq:dual_obj} $d(\valpha^\ast,\vv^\ast)$ coincides with the
minimum of \Eqref{eq:form_bpdn} $f(\vw^\ast)$, where $d$ is the objective
function in \Eqref{eq:dual_obj}; $\vw^\ast$ and
$(\valpha^\ast,\vv^\ast)$ are the minimizer and the maximizer of the
primal and dual problems, respectively.


The {\em augmented Lagrangian (AL)
function} of the dual problem (\Eqsref{eq:dual_obj} and
\eqref{eq:dual_const}) is defined as follows:
\begin{align}
\label{eq:dal_alf}
 L_\eta(\valpha,\vv;\vw)=
 -\frac{1}{2}\|\valpha-\vb\|_2^2+\frac{1}{2}\|\vb\|_2^2 -
 \delta_\lambda^\infty(\vv) -\vw\T\left(\mA\T\valpha-\vv\right) -\frac{\eta}{2}\|\mA\T\valpha-\vv\|_2^2, 
\end{align}
where $\vw$ is the Lagrangian multiplier associated with the equality
constraint (\Eqref{eq:dual_const}) and corresponds to the coefficient
vector in the primal problem. The last term in \Eqref{eq:dal_alf} is
called the barrier term and $\eta(\geq 0)$ is called the barrier
parameter. When $\eta=0$, the AL function is reduced to
the ordinary Lagrangian function. See \cite{Ber82,NocWri99} for the
details of the AL method. See also \cite{BoydBook} for the ordinary
Lagrangian duality. Now we define the surrogate function $f_\eta(\vw)$
as follows:
\begin{align}
\label{eq:fsurr}
 f_\eta(\vw)=\max_{\valpha\in\RR^m,\vv\in\RR^n}L_\eta(\valpha,\vv;\vw).
\end{align}
Note that from the strong duality
$f_0(\vw)=\max_{\valpha,\vv}L_0(\valpha,\vv;\vw)=f(\vw)$. In addition, since
$L_0(\valpha,\vv;\vw)\geq L_\eta(\valpha,\vv;\vw)$, the inequality
$f(\vw)\geq f_\eta(\vw)$ holds. Moreover, since
$f_\eta(\vw)\geq
L_\eta(\valpha^\ast,\vv^\ast;\vw)=d(\valpha^\ast,\vv^\ast)=f(\vw^\ast)$
(we use $\mA\T\valpha^\ast=\vv^\ast$ to obtain the first equality), we have
$\min_{\vw\in\RR^n} f_\eta(\vw)=f(\vw^\ast)$ for any nonnegative
$\eta$. Furthermore, $f_\eta(\vw)$ is differentiable if $\eta>0$.

The maximization with respect to $\vv$ in \Eqref{eq:fsurr}
can be computed analytically and
$\vv$ can be eliminated from \Eqref{eq:dal_alf}, as follows:
\begin{align*}
\max_{\vv\in\RR^n} L_\eta(\valpha,\vv;\vw)
 &=-\frac{1}{2}\|\valpha-\vb\|_2^2-\min_{\vv\in\RR^n}\left(
 \delta_\lambda^{\infty}(\vv)+\frac{\eta}{2}\left\|\vv-\mA\T\valpha-\vw/\eta\right\|_2^2\right)
 + c(\vw,\eta)\\
&=-\frac{1}{2}\|\valpha-\vb\|_2^2-\frac{\eta}{2}\|\mA\T\valpha+\vw/\eta-
 P^\infty_\lambda(\mA\T\valpha+\vw/\eta)\|_2^2+
 c(\vw,\eta)\\
&=-\frac{1}{2}\|\valpha-\vb\|_2^2-\frac{\eta}{2}\|{\sf
 ST}_{\lambda}(\mA\T\valpha+\vw/\eta)\|_2^2+c(\vw,\eta)=:L_\eta(\valpha;\vw),
\end{align*}
where $c(\vw,\eta)$ is a constant that only depends on $\vw$ and $\eta$,
and  $P^\infty_\lambda$ is a projection on the $\ell_\infty$ ball of
radius $\lambda$; note that $\eta P_\lambda^\infty(\vw)=P_{\eta
\lambda}^\infty(\eta\vw)$;  
in addition,
we define the well known {\em soft thresholding function} ${\sf ST}_{\lambda}$ (see
\cite{FigNow03,DauDefMol04,ComWaj05,YinOshGolDar08}) as follows:
\begin{align*}
 {\sf ST}_\lambda(\vw)&=\vw - P_\lambda^\infty(\vw)= 
\begin{cases}
 w_j-\lambda & \textrm{if $w_j>\lambda$},\\
 0           & \textrm{if $-\lambda\leq w_j\leq \lambda$},\\
 w_j+\lambda & \textrm{if $w_j<-\lambda$},
\end{cases}\quad(j=1,\ldots,n).
\end{align*}

\begin{figure}[tbp]
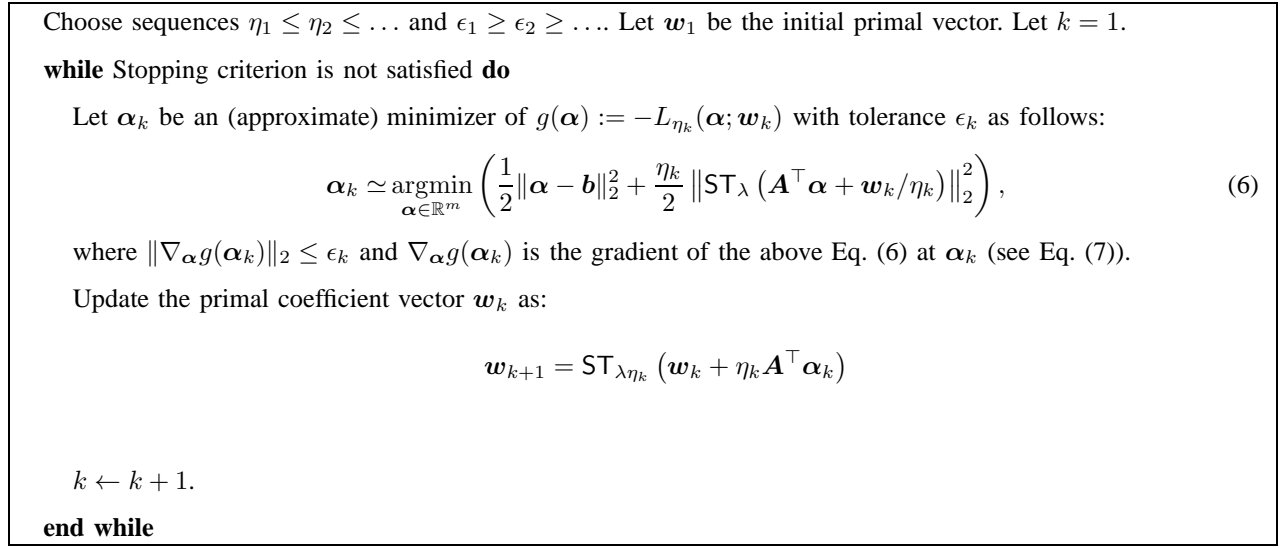

 \fbox{\small
\begin{minipage}[cbt]{\columnwidth}
 \begin{algorithmic}
  \STATE{Choose sequences $\eta_1\leq\eta_2\leq\ldots$ and
       $\epsilon_1\geq\epsilon_2\geq\ldots$. Let $\vw_1$ be the initial
       primal vector. Let $k=1$.}
  \WHILE{Stopping criterion is not satisfied}
  \STATE{Let $\valpha_k$ be an (approximate) minimizer of
       $g(\valpha):=-L_{\eta_k}(\valpha;\vw_k)$ with tolerance $\epsilon_k$ as
       follows:
\begin{align}
\label{eq:dal_obj}
 \valpha_k\simeq&\argmin_{\valpha\in\RR^m}\left(\frac{1}{2}\|\valpha-\vb\|_2^2+\frac{\eta_k}{2}\left\|{\sf
 ST}_{\lambda}\left(\mA\T\valpha+\vw_k/\eta_k\right)\right\|_2^2\right),
\end{align}
 where $\|\nabla_{\valpha} g(\valpha_k)\|_2\leq
 \epsilon_k$ and $\nabla_{\valpha} g(\valpha_k)$ is the gradient of the above
\Eqref{eq:dal_obj}   at $\valpha_k$ (see \Eqref{eq:dal_grad}).}
  \STATE{Update the primal coefficient vector $\vw_k$ as:
\begin{align*}
 \vw_{k+1}={\sf ST}_{\lambda\eta_k}\left(\vw_k+\eta_k\mA\T\valpha_k\right)
\end{align*}}
  \STATE{$k\leftarrow k+1$.}
  \ENDWHILE
 \end{algorithmic}
\end{minipage}
}
\caption{Dual augmented Lagrangian method for sparse signal
 reconstruction (see \Eqsref{eq:form_bpdn} and \eqref{eq:dual_obj}.)}
\label{fig:dal_algo}
\end{figure}

Typically in an AL method the barrier parameter $\eta$ is increased as
$\eta_1\leq\eta_2\leq\ldots$; this guarantees super linear convergence
of the method (see~\cite{Ber82}). The coefficient vector $\vw$ is updated using the
gradient of $f_\eta(\vw)$ as follows:
\begin{align*}
\vw_{k+1} = \vw_{k} + \eta_k(\mA\T\valpha_k-\vv_k)
=\vw_{k}+\eta_k(\mA\T\valpha_k-P_\lambda^\infty(\mA\T\valpha_k+\vw_k/\eta_k))={\sf
 ST}_{\lambda\eta_k}(\vw_k+\eta_k\mA\T\valpha_k),
\end{align*}
because $\nabla_{\vw}
f_\eta(\vw_k)=(\nabla_{\vw}\valpha_k)\nabla_{\valpha}
L_\eta(\valpha_k;\vw_k)+\nabla_{\vw}L_\eta(\valpha_k;\vw_k)=-(\mA\T\valpha_k-\vv_k)$, 
where $\valpha_k$ and $\vv_k$ are the maximizer of \Eqref{eq:fsurr} at
the current $\vw_k$ and $\nabla_{\valpha}L_\eta(\valpha_k;\vw_k)=\bm{0}$
because $\valpha_k$ maximizes $L_\eta(\valpha;\vw_k)$. We can also show that $f(\vw_k)\geq
f_\eta(\vw_k)\geq f(\vw_{k+1})$ with strict inequality except the
minimum of \Eqref{eq:form_bpdn}~\cite[Chap.5]{Ber82}. Accordingly the dual augmented Lagrangian method can be described as in \Figref{fig:dal_algo}.

\subsection{Inner minimization}

Let $g(\valpha)$ be the
objective function in \Eqref{eq:dal_obj}; $g(\valpha)$ is once differentiable
everywhere and also twice differentiable except the points on which the
above soft thresholding function switches. We use the Newton method for
the minimization of $g(\valpha)$.  The gradient and the Hessian
of the objective function $g(\valpha)$  can be written as follows:
\begin{align}
\label{eq:dal_grad}
 \nabla_{\valpha} g(\valpha)&=\valpha-\vb+\eta_k\mA{\sf
 ST}_\lambda(\vq),\\
\label{eq:dal_hess}
 \nabla_{\valpha}^2 g(\valpha)&=\mI_m+\eta_k\mA_+\mA_+\T,
\end{align}
where $\vq=\mA\T\valpha+\vw_k/\eta_k$, $\mI_m$ is the identity matrix of size
$m$, and $\mA_+$ is the submatrix
of $\mA$ that consists of ``active'' columns with indices
$\mathcal{J}_+=\{j\in\{1,2,\ldots,n\}:|q_j|>\lambda\}$.
Note that in both the computation of the gradient and the Hessian,
computational complexity is only proportional to the number of active components of $\vq$.
The discontinuity of the second derivative is in general not a
problem. In fact, we can see from the complementary slackness condition
that for finite $\eta$ the optimal solution-multiplier pair
$(\vw^\ast,\valpha^\ast)$ is on a regular point; thus the convergence
around the minimum is quadratic. 

We propose two approaches for 
solving the Newton system
$\nabla_{\valpha}^2g(\valpha)\vy=-\nabla_{\valpha} g(\valpha)$.
The first approach (DALchol) uses the Cholesky factorization of the Hessian matrix
$\nabla_{\valpha}^2g(\valpha)$. The second approach (DALcg) uses a
preconditioned conjugate gradient method (the truncated Newton method
in~\cite{KimKohLusBoyGor07}) with a preconditioner that only consists of
the diagonal elements of the Hessian matrix. Finally the standard backtracking
line-search with initial step-size $1$ is applied to guarantee decrease
in the objective $g(\valpha)$.

\section{Empirical comparisons}
\label{sec:results}
We test the computational efficiency of the proposed DAL algorithm on 
the $\ell_2$-$\ell_1$ problem (\Eqref{eq:form_bpdn}) under various
conditions. The DAL algorithm is compared to two state-of-the-art
algorithms, namely l1\_ls (interior-point
algorithm,~\cite{KimKohLusBoyGor07}) and SpaRSA (step-size improved IST,~\cite{WriNowFig09}).

\subsection{Experimental settings}
In the first experiment (\Figref{fig:normal}), the elements of the design
matrix $\mA$ are sampled from the independent zero-mean Gaussian
distribution with variance $1/(2n)$. This choice of variance makes the
largest singularvalue of $\mA$ approximately one~\cite{WriNowFig09}. The true coefficient
vector $\vw_0$ is generated by randomly filling $4\%$ of its
elements by $+1$ or $-1$ which is also randomly chosen. The remaining
elements are zero. The target vector $\vb$ is generated as
$\vb=\mA\vw_0+\vxi$, where $\vxi$ is sampled from the zero-mean Gaussian
distribution with variance $10^{-4}$. The number of observations ($m$)
is increased from $m=128$ to $m=8,196$ while the number of variables
($n$) is increased proportionally as $n=4m$.
The regularization constant $\lambda$ is kept constant at $0.025$, which
is found to approximately correspond to the choice
$\lambda=0.1\|\mA\T\vb\|_\infty$ in
\cite{WriNowFig09}\footnote{A fair comparison
at a smaller regularization constant would require continuation techniques,
which should be addressed in a separate paper.}.
In the second experiment (\Figref{fig:poor}), the setting is almost the
same except that the singular values of $\mA$ is replaced by a series
decreasing as $1/s$ for the $s$-th singular value. Thus the condition
number (the ratio between the smallest and the largest singular values) of $\mA$ is $m$. Additionally we set the variance of $\vxi$ to
zero (no noise) and keep $\lambda$ constant at $0.0003$, which is also
found to approximately correspond to the setting in \cite{WriNowFig09}
In the last experiment (\Figref{fig:largescale}), the number of
observations ($m$) is kept at $m=1,024$ and the number of samples ($n$)
is increased from $n=4,096$ to $n=1,048,576$. The design matrix $\mA$
and the target vector $\vb$ are generated as in the first
experiment. In addition, the regularization constant $\lambda$ is
decreased as $\lambda=1.6/n^{1/2}$, which equals $0.025$ at $n=4,096$
and is again chosen to approximately match the setting in \cite{WriNowFig09}.
In each figure, we show the computation time, the number of steps, and the
sparsity of the solution (the proportion of non-zero elements in the
final solution) from top to bottom. All the results are averaged over
10 random initial coefficient vectors $\vw$. 
All the experiments are run on MATLAB~7.7 (R2008b) on a workstation with
two 3.0GHz quad-core Xeon processors and 16GB of memory.

\subsection{Practical issues}
\subsubsection{Stopping criterion}
We use the ``duality'' stopping criterion proposed in
\cite{WriNowFig09} for all the results presented
in the next section. More precisely, we generate a dual variable
$\hat{\valpha}$ as follows,
\begin{align*}
 \hat{\valpha} = \lambda{\tilde{\valpha}}/{\|\mA\T\tilde{\valpha}\|_\infty},
\end{align*}
where $\tilde{\valpha}=\mA\vw-\vb$ is the gradient of the
primal loss term in \Eqref{eq:form_bpdn}. The above defined
$\hat{\valpha}$ is a feasible point of the dual problem (\Eqref{eq:dual_obj}) by
definition, i.e., $\|\mA\T\hat{\valpha}\|_\infty\leq\lambda$. Thus we use
the primal-dual pair $(\vw,\hat{\valpha})$ to measure the relative
duality gap $(f(\vw)-d(\hat{\valpha},\mA\T\hat{\valpha}))/f(\vw)$, where $f$ and $d$ are
the objective functions in the primal problem (\Eqref{eq:form_bpdn}) and
the dual problem (\Eqref{eq:dual_obj}), respectively. The tolerance $10^{-3}$ is used.
\subsubsection{Hyperparameters}
The tolerance parameter $\epsilon_k$ for the inner minimization is
chosen as follows. We use $\epsilon_1= 10^{-4}\cdot{m}^{1/2}$ and decrease $\epsilon_k$ as
$\epsilon_k=\epsilon_{k-1}/2$. Using larger $\epsilon_k$ results in
cheaper inner minimization but often requires a larger number of outer
iterations. The barrier
parameter $\eta_k$ also affects the behavior of the
algorithm. Typically larger $\eta_k$ gives larger reduction in the
duality gap at every iteration but makes the inner minimization more
difficult. Additionally the best value of $\eta_k$ depends on the size
of the problem, regularization constant $\lambda$, and the spectrum of
$\mA$. We manually choose $\eta_1$ for each problem in the next section
and increase $\eta_k$ as $\eta_k=2\eta_{k-1}$, which guarantees
the super-linear convergence~\cite{Ber82}.

\subsection{Results}
\begin{figure}
 \begin{center}
  \subfigure[Normal
  conditioning]{\includegraphics[width=0.5\columnwidth]{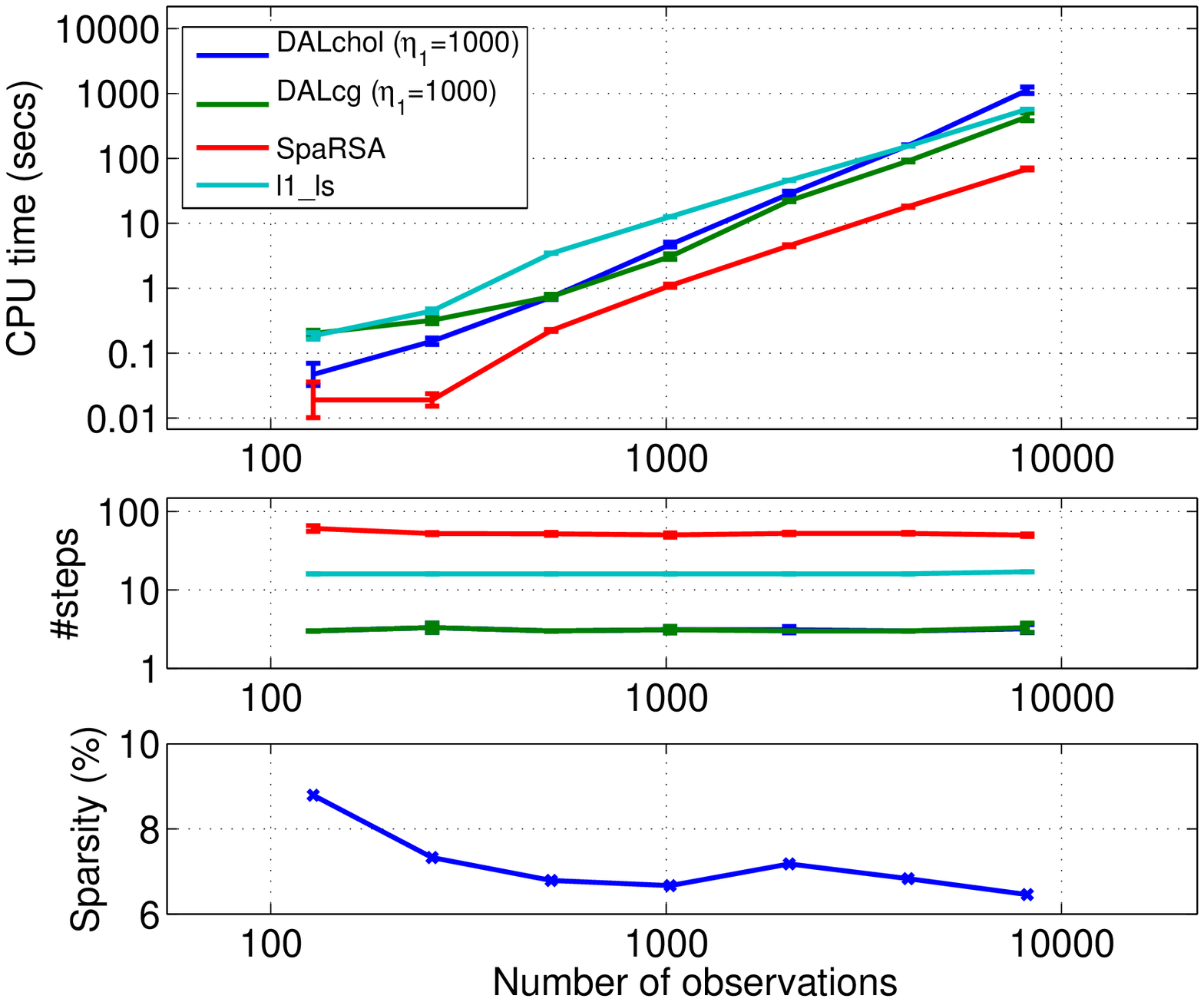}\label{fig:normal}}~\subfigure[Poor
conditioning]{\includegraphics[width=0.5\columnwidth]{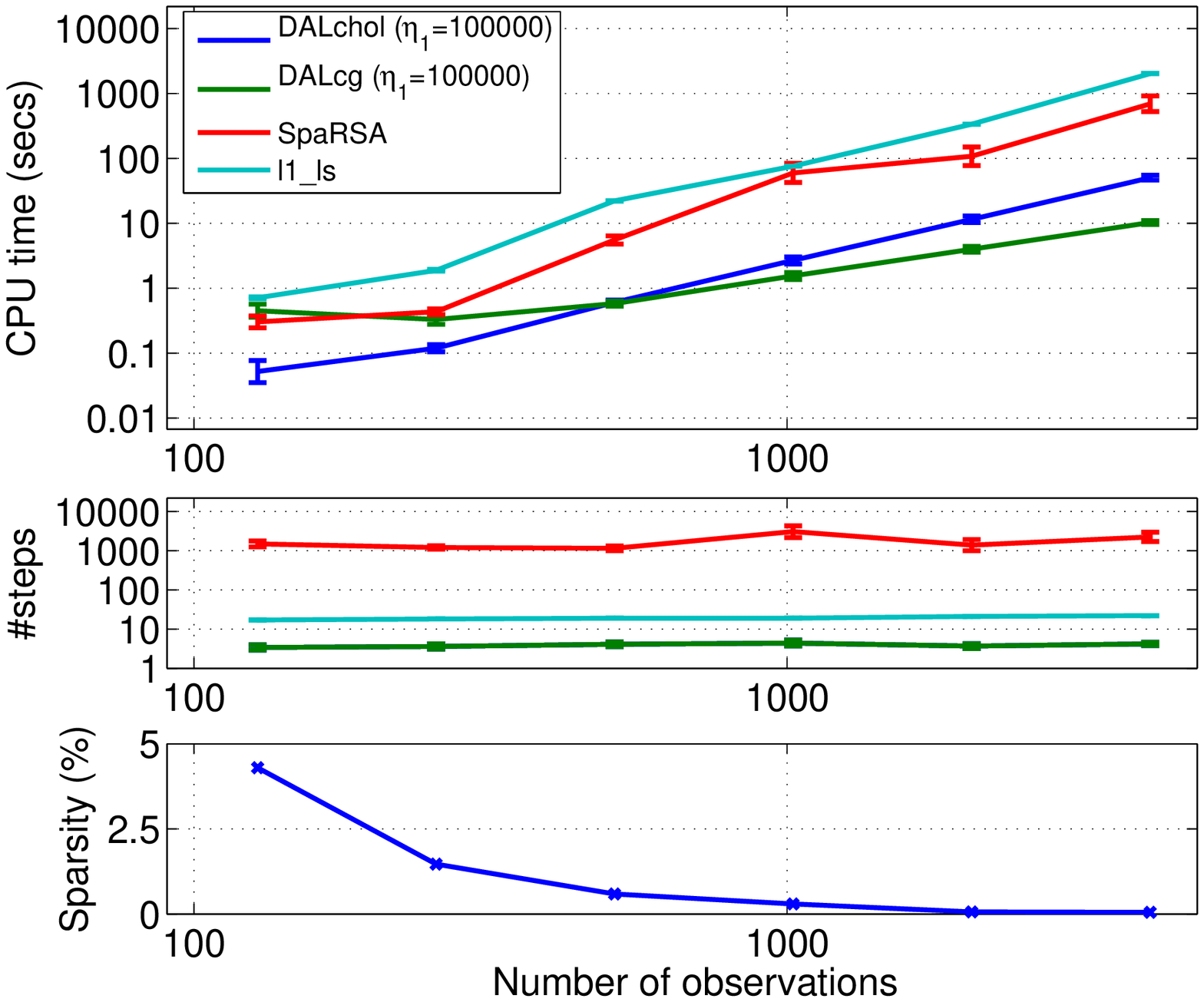}\label{fig:poor}}
  \caption{Comparison of running time and number of steps for three
  optimization algorithms, namely DAL, SpaRSA and l1\_ls for problems of
  various sizes with (a) design matrix $\mA$ generated from independent
  normal random variables and (b) the same matrix with singular values
  replaced by a power-law distribution. The horizontal axis denotes the
  number of observations ($m$). The number of variables is $n=4m$.
 The regularization constant
  $\lambda$ is fixed at $\lambda=0.025$ in (a) and $\lambda=0.0003$ in
  (b). }
  \label{fig:scaling}
 \end{center}
\end{figure}

When the data is well conditioned (\Figref{fig:normal}), SpaRSA
performs clearly the best within the three algorithms. The proposed DAL
algorithm with the conjugate gradient (DALcg) performs comparable to
l1\_ls. The proposed DAL with the Cholesky factorization (DALchol) is less
efficient than DALcg when $m$ is large because
the complexity grows as $O(m^3)$; note however that the cost for building
the Hessian matrix is only $O(m^2n_+)$, where $n_+$ is the number of
active components (see \Eqref{eq:dal_hess}).

In contrast, when the data is poorly conditioned (\Figref{fig:poor}),
the proposed DALcg runs almost $100$ times faster than SpaRSA at most. 
This can be clearly seen in the number of steps (the middle row). Although
the numbers of steps DAL and l1\_ls require are almost constant from
\Figref{fig:normal} to \Figref{fig:poor}, that of SpaRSA is increased at
least by the factor $10$. Note that the sparsity of the solution is
decreasing as 
the number of samples increases. This may explain why the proposed
DAL algorithm is more robust to poor conditioning than l1\_ls because
l1\_ls does not exploit the sparsity in the solution. 

\begin{figure}[tbp]
 \begin{center}
  \includegraphics[width=.7\columnwidth]{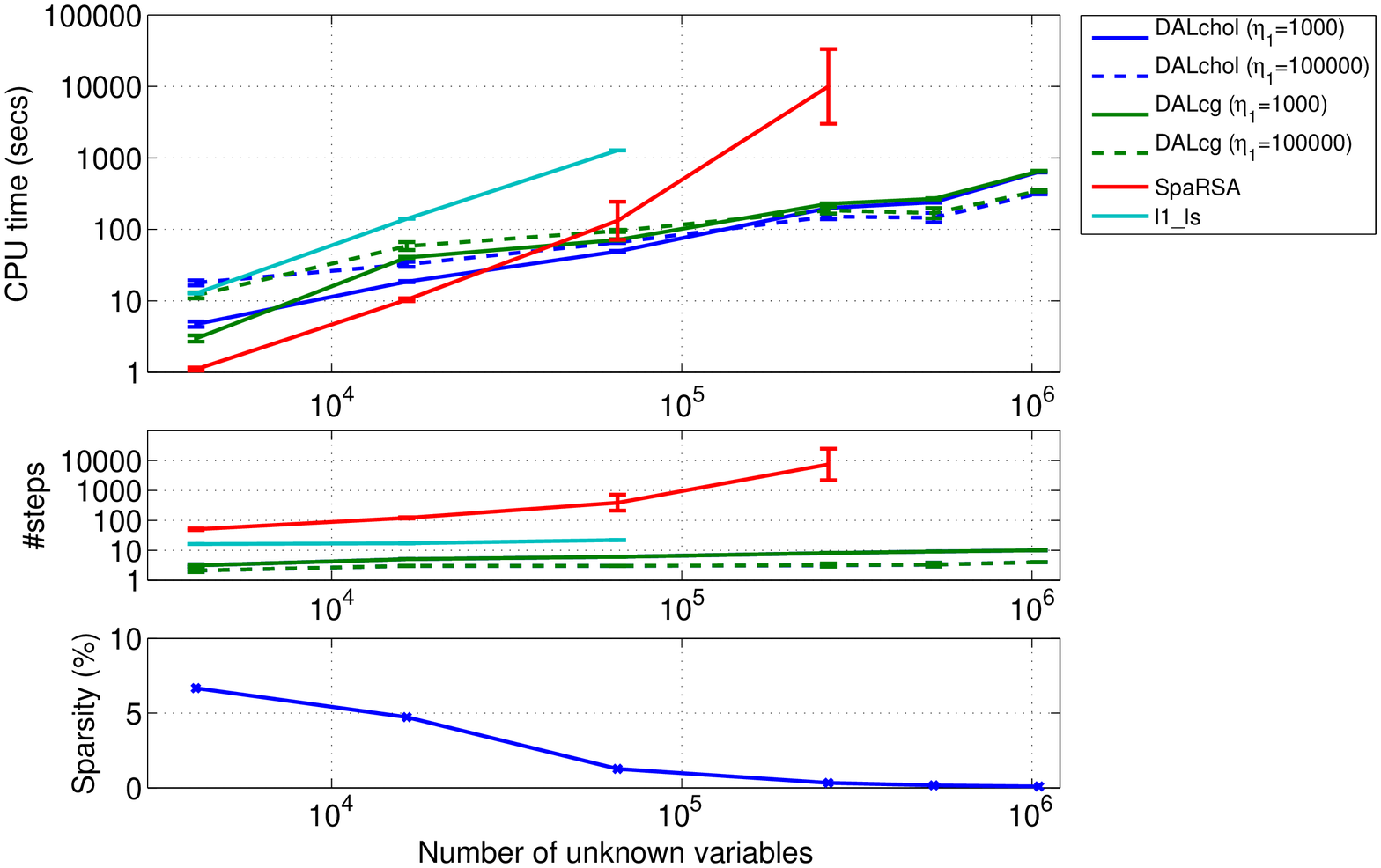}
  \caption{Comparison of the algorithms for large scale problems when
  the number of variable ($n$) is much larger than the number of
  observations ($m$). $m$ is kept constant at $m=1024$. $\lambda$ is
  decreased as $\lambda=1.6/n^{1/2}$.}
  \label{fig:largescale}
 \end{center}
\end{figure}

Finally we compare the three algorithms for very large problems
in~\Figref{fig:largescale}. Clearly the proposed DAL has milder
scaling to the dimensionality than both SpaRSA and l1\_ls. This is
because the proposed DAL algorithm is based on the dual problem
(\Eqref{eq:dual_obj}). The computational efficiency of DALchol and DALcg
is comparable because $m$ is kept constant in this experiment. The
initial barrier parameter $\eta_1=100000$ seems to perform better than
$\eta_1=1000$ for large $n$.


\section{Conclusion}
\label{sec:summary}
In this paper we have proposed a new optimization framework for sparse signal
reconstruction, which converges super-linearly. It is based on the dual sparse reconstruction
problem. 
The sparsity of the coefficient vector $\vw$ is explicitly used in the
algorithm. 
%
Numerical comparisons have shown that the proposed DAL algorithm is
favorable against a state-of-the-art algorithm SpaRSA when the design
matrix $\mA$ is poorly conditioned or $m\ll n$. 
In fact, it has solved problems with  millions of
variables in less than 20 minutes even when the design matrix $\mA$ is dense. 
In addition, for dense
$\mA$, DAL has shown improved efficiency to l1\_ls in most cases.
Future work includes generalization of DAL to other loss functions
and sparsity measures, continuation strategy,
and approximate minimization of the inner problem.

\section*{Acknowledgment}
The authors would like to thank Masakazu Kojima, David Wipf, Srikantan
Nagarajan, and Hagai Attias for helpful discussions. RT was supported by the Global COE program
(Computationism as a Foundation for the Sciences). 
MS is supported by
MEXT Grant-in-Aid for Young Scientists (A) 20680007,
SCAT, and AOARD. This work was done while RT was at Tokyo Institute of Technology.




%
\bibliographystyle{IEEEtran}
\bibliography{IEEEabrv,TomSug09}

\end{document}